\documentclass[review,authoryear,jmlmc]{beg_32}      %  review version
\usepackage{hyperref}
\usepackage[normalem]{ulem}
\usepackage[hang]{footmisc}
\fancypagestyle{plain}{%
  \fancyhf{}
  \fancyhead[R]{\small {\it \jname}, x(x):\thepage--\pageref{LastPage} (\myyear\today)}
  \fancyfoot[R]{\small\bf\thepage }
  \fancyfoot[L]{\fottitle}
  }

\renewcommand{\dmyy}{20}
\renewcommand{\myyear}{2022}
\renewcommand{\today}{}
\begin{document}

\volume{Volume x, Issue x, (pre-print submitted to JMLMC) \myyear\today}
\title{GeoThermalCloud:~Machine Learning for Geothermal Resource Exploration}
\titlehead{GeoThermalCloud: Machine Learning for Geothermal Resource Exploration}
\authorhead{M.~K.~Mudunuru, B.~Ahmmed, and V.~V.~Vesselinov}
%For at least  authors with different addresses, use instead the following commands
\corrauthor[1]{Maruti K. Mudunuru}
\author[2]{Velimir V. Vesselinov}
\author[3]{Bulbul Ahmmed}
\corremail{maruti@pnnl.gov}
\corraddress{Earth System Measurement \& Data Group, Atmospheric Sciences \& Global Change Division, Pacific Northwest National Laboratory, Richland, WA 99352, USA. Email:~maruti@pnnl.gov}
\address[1]{Earth System Measurement \& Data Group, Atmospheric Sciences \& Global Change Division, Pacific Northwest National Laboratory, Richland, WA 99352, USA. Email:~maruti@pnnl.gov}
\address[2]{EnviTrace LLC. Santa Fe, NM 87501, USA. Email:~velimir.vesselinov@gmail.com}
\address[3]{Earth and Environmental Sciences Division, Los Alamos National Laboratory, Los Alamos, NM 87545, USA. Email:~ahmmedb@lanl.gov}
% \address[2]{Business or Academic Affiliation 2, City, Province, Zip Code, Country Business or Academic Affiliation 2, City, Province, Zip Code, Country}
% End information for at least  authors with different addresses
% For authors with the same post address,
%\corrauthor{First A. Author}
%\corremail{f.author@affiliation.com}
%\author{Second B. Author, Jr.}
%\address{Department of Chemistry and Courant, Institute of Mathematical Sciences, New York, NY 10012, USA}
% End commands for all authors with the same address

\dataO{mm/dd/yyyy}
%\dataO{}
\dataF{mm/dd/yyyy}
%\dataF{}

\abstract{}


\maketitle

\graphicspath{{Figures/}}

%%%%%%%%%%%%%%%%%%%%%%%%%%%%%%%%%%%%%%%%%%%%%%%%%
\section*{Abstract}
%%%%%%%%%%%%%%%%%%%%%%%%%%%%%%%%%%%%%%%%%%%%%%%%%
{\scriptsize
Geothermal is a renewable energy source that can provide reliable and flexible electricity generation for the world.
In the past decade, the U.S. Geological Survey's resource assessments, Play Fairway Analyses (PFA), and GeoVision report by the U.S. Department of Energy's Geothermal Technologies Office provided insights on enormous untapped potential for geothermal energy to contribute to the U.S. domestic energy needs.
Past studies identified that geothermal resources without surface expression (e.g., blind/hidden hydrothermal systems) have vast potential.
These blind systems can significantly increase power generation.
But a primary challenge is locating and quantifying these hidden resources, which do not have any thermal manifestations on the surface.
PFA has successfully identified some blind systems in the western USA (e.g., specific locations in the Great Basin region within Nevada).
However, a comprehensive search for these blind systems can be time-consuming, expensive, and resource-intensive, with a low probability of success.
Accelerated discovery of these blind resources is needed with growing energy needs and higher chances of exploration success.
Recent advances in machine learning (ML) have shown promise in shortening the timeline for this discovery.
This paper presents a novel ML-based methodology for geothermal exploration towards PFA applications.
Our methodology is provided through our open-source ML framework, GeoThermalCloud \url{https://github.com/SmartTensors/GeoThermalCloud.jl}.
The GeoThermalCloud uses a series of unsupervised, supervised, and physics-informed ML methods available in SmartTensors AI platform \url{https://github.com/SmartTensors}.
Here, the presented analyses are performed using our unsupervised ML algorithm called NMF$k$, which is available in the SmartTensors AI platform.
Our ML algorithm facilitates the discovery of new phenomena, hidden patterns, and mechanisms that helps us to make informed decisions.
Moreover, the GeoThermalCloud enhances the collected PFA data and discovers signatures representative of geothermal resources.
Through GeoThermalCloud, we could identify hidden patterns in the geothermal field data needed to discover blind systems efficiently.
Crucial geothermal signatures often overlooked in traditional PFA are extracted using the GeoThermalCloud and analyzed by the subject matter experts to provide ML-enhanced PFA, which is informative for efficient exploration.
We applied our ML methodology to various open-source geothermal datasets within the U.S. (some of these are collected by past PFA work). The results provide valuable insights into resource types within those regions.
This ML-enhanced workflow makes the GeoThermalCloud attractive for the geothermal community to improve existing datasets and extract valuable information often unnoticed during geothermal exploration.
}

%%%%%%%%%%%%%%%%%%%%%%%%%%%%%%%%%%%%%%%%%%%%%%%%%
\section{Introduction}
\label{sec:intro}
%%%%%%%%%%%%%%%%%%%%%%%%%%%%%%%%%%%%%%%%%%%%%%%%%
Geothermal energy is a nearly inexhaustible renewable resource that harnesses Earth's heat \cite{brown2012mining}.
Broad deployment of geothermal energy can provide essential benefits to the U.S. domestic energy needs, including electricity production, thermal energy storage, mining of critical materials (e.g., lithium), and efficient heating and cooling solutions \cite{wicks2022new,cheng2022president}.
As of 2021, the U.S. installed geothermal powerplant capacity is more than 3.8 GigaWatts (GW) \cite{hamm2021geothermal}.
In the last decade, the U.S. Department of Energy (DOE) Geothermal Technologies Office (GTO) recently conducted a study called the GeoVision to understand and assess the geothermal potential in our country \cite{GeoVi2019}.
Based on the state-of-the-art geothermal technology (e.g., recent advances in drilling and data analytics), the 2019 GeoVision analysis report suggests that more than 30 GW of new and undiscovered hydrothermal resources is recoverable.
Additionally, we can access 100+ GW of new geothermal energy by developing enhanced geothermal systems (EGS).
However, various technical and non-technical barriers exist to successful exploration, development, and widespread geothermal energy deployment \cite{imolauer2010non,levine2017crossing,young2017crossing}.
A significant barrier is efficient discovery and accelerated exploration of hidden or blind geothermal resources \cite{dobson2016review}.

Blind systems are shown to have three times more resource potential within the same region than existing and identified hydrothermal systems \cite{dobson2016review}.
However, these systems lack surface thermal expressions at the ground surface, making it more challenging to discover them compared to traditional, surface-manifested hydrothermal resources.
As a result, the probability of successfully locating and developing these resources without a comprehensive and expensive exploration and research is very low \cite{gehringer2012geothermal}.
To overcome these challenges of discovering blind systems, the U.S. DOE-GTO has funded multiple projects through three phases to develop a systematic approach to quantify geothermal prospects while reducing overall exploration costs \cite{hamm2021geothermal}.
This approach to improving geothermal development projects' success rate is called Play Fairway Analysis (PFA).
The PFA concept borrowed from oil\& gas industries allows us to identify potential locations of blind hydrothermal systems and quantify geothermal potential in the regions of interest \cite{holmes2022machine}.
Traditional PFA was able to identify various regions of interest for geothermal exploration \cite{faulds2016nevada}.
However, the main limitation of this traditional analysis is effectively combining sparse and multi-source data streams with missing values to explore geothermal resources better \cite{vesselinov2020discovering,mudunuru2020site,siler2021machine,smith2021machine,moraga2022geothermal}.

In this study, we present GeoThermalCloud, a novel ML methodology to enhance and accelerate PFA by overcoming these limitations.
Unsupervised (self-supervised) ML algorithms based on non-negative matrix factorization and custom $k$-means clustering (NMF$k$) provide means to overcome this challenge \cite{alexandrov2014blind,vesselinov2019unsupervised}.
The NMF$k$ algorithm is one of the tools available in the SmartTensors AI platform \url{https://github.com/SmartTensors} to preprocess sparse datasets with missing values.
The GeoThermalCloud is developed based on this powerful AI platform for extracting hidden geothermal signatures from field data.
Specifically, the GeothermalCloud facilitates curation and analysis of geothermal datasets \url{https://github.com/SmartTensors/GeoThermalCloud.jl}.
The uniqueness of the GeoThermalCloud is that it is flexible, open-source, and provides site/regional data, scripts, examples, figures, and developed geothermal prospectivity maps for the geothermal community.
It has in-build preprocessing, postprocessing, and state-of-the-art visualization tools for non-experts.
Therefore, experts and non-experts can equally utilize this capability without going through a steep learning curve.
Based on subject matter expertise (SME), the discovered signatures using the GeoThermalCloud are assessed and correlated to geothermal resource type (e.g., low, medium, and high-temperature resources) \cite{siler2021machine}.
We can use the extracted signatures upon SME assessment to build geothermal prospectivity, develop exploration risk maps, and guide future data collection strategies \cite{vesselinov2022geothermalcloud}.

The paper is organized as follows:~Section~\ref{sec:intro} discusses state-of-the-art for PFA and its limitations.
Section~\ref{sec:Methods} describes the proposed methodology for ML-enhanced PFA.
Section~\ref{sec:data} describes the PFA data curated from open-source geothermal repositories for NMF$k$ analysis.
Sec.~\ref{sec:results} presents the results and discussion of ML-analyzed geothermal data for popular PFAs in the USA.
Finally, Sec.~\ref{sec:Conclusions} presents our conclusions.

%%%%%%%%%%%%%%%%%%%%%%%%%%%%%%%%%%%%%%%%%%%%%%%%%
\section{Proposed methodology}
\label{sec:Methods}
%%%%%%%%%%%%%%%%%%%%%%%%%%%%%%%%%%%%%%%%%%%%%%%%%
This section describes the proposed ML methodology used to build the GeoThermalCloud.
GeoThermalCloud capabilities and workflow include (1) analyzing large field datasets, (2) assimilating model simulations (large inputs and outputs), (3) processing sparse datasets, (4) performing transfer learning (between sites with different exploratory levels), (5) extracting hidden geothermal signatures in the field and simulation data, (6) labeling geothermal resources and processes, (7) identifying high-value data acquisition targets, and (8) guiding geothermal exploration and production by selecting optimal exploration, production, and drilling strategies.
In Traditional PFA analysis, first, the SMEs review and analyze the data to estimate the geothermal potential in a region \cite{faulds2016nevada,lautze2017play,siler2017play,ito2017play,lindsey2021play}.
Second, dataset attributes and associated evidence of geothermal system occurrence are combined using weighted products or weighted sums.
Finally, geothermal PFA models are generated to estimate the site- or regional-scale prospectivity in the form of maps or other data visualization tools.
Our ML-enhanced PFA workflow (ePFA) has four steps combining the strengths of SMEs and ML to develop maps of geothermal potential \cite{ahmmed2022machine}.
The site- or regional-scale data are curated and processed using ML to discover hidden features/signatures in the first step.
SMEs interpret these signatures to associate them with a resource type (e.g., low/medium/high-temperature geothermal systems).
In the second step, ML analyses are performed at other sites/regions to transfer knowledge (e.g., reusing discovered geothermal patterns, signatures, and relationships between geothermal data attributes) to the analyzed site/regional datasets.
In the third step, physics insights gained from process models and simulations are applied to enrich the analyzed dataset (e.g., fill data gaps).
Finally, the information from the above three individual steps is combined using risk analysis and uncertainty quantification methods to develop site-specific prospectivity maps and future data acquisition strategies.

For this ePFA workflow development, we use an unsupervised ML algorithm called NMF$k$ \cite{alexandrov2014blind,VESSELINOV2022102576} to discover hidden patterns in geothermal datasets.
The NMF$k$ is a machine learning method that combines non-negative matrix factorization (NMF) with customized $k$-means clustering.
NMF learns a parts-based representation \cite{Lee1999,cichocki2009} of the analyzed geothermal data matrix.
The $k$-means clustering groups similar NMF learned pieces together and discovers unique underlying patterns \cite{wagstaff2001kmeans}.
Customization of the $k$-means clustering provides insights on the similarity of learned parts within a given cluster compared to other identified groups \cite{iliev2018nonnegative,vesselinov2019unsupervised}.
This study builds a data matrix, $X$ of size ($n$, $m$), after curating the raw geothermal datasets.
This geothermal data matrix is developed by combining geophysical, geological, hydrological, and thermal data sources.
For instance, the DOE's Geothermal Data Repository (\url{https://gdr.openei.org/}) provides a venue to download such datasets for ML analysis.
The preprocessor normalizes the raw data such that the elements of $X$ are non-negative.
If needed, it also log transforms individual attributes to reduce the effects of outliers and nonnormal distributions.
Here, $n$ is the number of geothermal data attributes, and $m$ is the number of spatial locations of geothermal measurements.
The NMF$k$ is part of SmartTensors AI Platform and can be found at \url{https://github.com/SmartTensors}.

NMF of NMF$k$ algorithm decomposes $X_{n \times m}$ into two matrices, $W_{n \times k}$ (basis/mixing matrix) and $H_{k \times m}$ (coefficient/attribute matrix) as:
\begin{equation}\label{eq:nmf1_factorize}
    X = W \times H + \epsilon(k),
\end{equation}
where $\epsilon(k)$ is an error matrix representing the discrepancy between $X$ and its estimate $W \times H$ for the specified $k$ while $k$ is the unknown/hidden number of signatures present in the data and is always smaller than $n$ and $m$.
The $W$ matrix represents how the measurement locations are related to the hidden signatures.
The $H$ matrix depicts the relationship between attributes and hidden signatures.
In this methodology, the number of hidden signatures ($k$) is unknown and is identified by performing a series of NMF analyses for $k = 2, 3,\cdots,d$, where $d \leq m$.
The NMF process minimizes the following objective function, $\mathcal{L}$, based on Frobenius norm for a specified $k$, such that the entries of the resulting $W$ and $H$ matrices are non-negative:
\begin{equation}\label{eq:NMF2_objective}
    \mathcal{L} = ||X - W \times H ||_F \quad \mathrm{such \; that} \quad W, H \geq 0 \quad \forall \quad n, m, k.
\end{equation}{}
In the minimization, the following rules are used to update the $W$ and $H$ matrices \cite{Lee1999}:

\begin{equation}\label{eq:NMF_W}
    W_{ij} \leftarrow W_{ij} \frac{(XH^T)_{ij}}{(WHH^T)_{ij}},
\end{equation}
and
\begin{equation}\label{eq:NMF_H}
    H_{ij} \leftarrow H_{ij} \frac{(W^TX)_{ij}}{(W^TWH)_{ij}}.
\end{equation}
For each $k$, NMF is solved for a series of random initial guesses (typically, 1,000 or more) for $W$ and $H$ matrices.
The least value of $\mathcal{L}$ for a given $k$ is assumed as the best value for the reconstruction error.
After completing the NMF process, the columns of the 1,000 estimated $H$ matrices are clustered into $k$ clusters using a customized $k$-means clustering.
Alternatively, we cluster the rows of the 1,000 estimated $W$ matrices.
Typically, we cluster the smaller matrix.
However, $k$ is also unknown in the $k$-means clustering.
The algorithm consecutively examines a specified $k$ by obtaining 1,000 $H$ matrices for each feature.
During clustering, the similarity between two clusters is assessed using the Silhouette width/value \cite{silhouettes}, which is essentially the cosine norm:
\begin{align}\label{eq:NMF3_silhoutte}
    \rho\left(p,q\right) = 1 - \frac{\sum_{i=1}^n p_i q_i}{\sum_{i=1}^n p_i^2 \sum_{i=1}^n q_i^2},
\end{align}
where $p_i$ and $q_i$ are components of vectors $\mathbf p$ and $\mathbf q$.
The Silhouette value quantifies how similar an object is to its own cluster compared to other clusters and varies from $-1$ to $+1$; high values indicate that the object is well matched to its own cluster and poorly matched to neighboring clusters.
The combination of reconstruction error ($\mathcal{L}(k)$) and the Silhouette value are used to determine the optimal number of hidden signatures.
If $k$ is low, the Silhouette value will be high, but so maybe $\mathcal{L}(k)$ because of under-fitting.
For high $k$, the Silhouette value will be low and the solution may be over-fit.
So, the best estimate for $k$ is a number that optimizes both $\mathcal{L}(k)$ and the Silhouette value.

%%%%%%%%%%%%%%%%%%%%%%%%%%%%%%%%%%%%%%%%%%%%%%%%%
\section{Data}
\label{sec:data}
\begin{figure}[htbp]
  \centering
    \includegraphics[width = 0.5\textwidth,page=1]{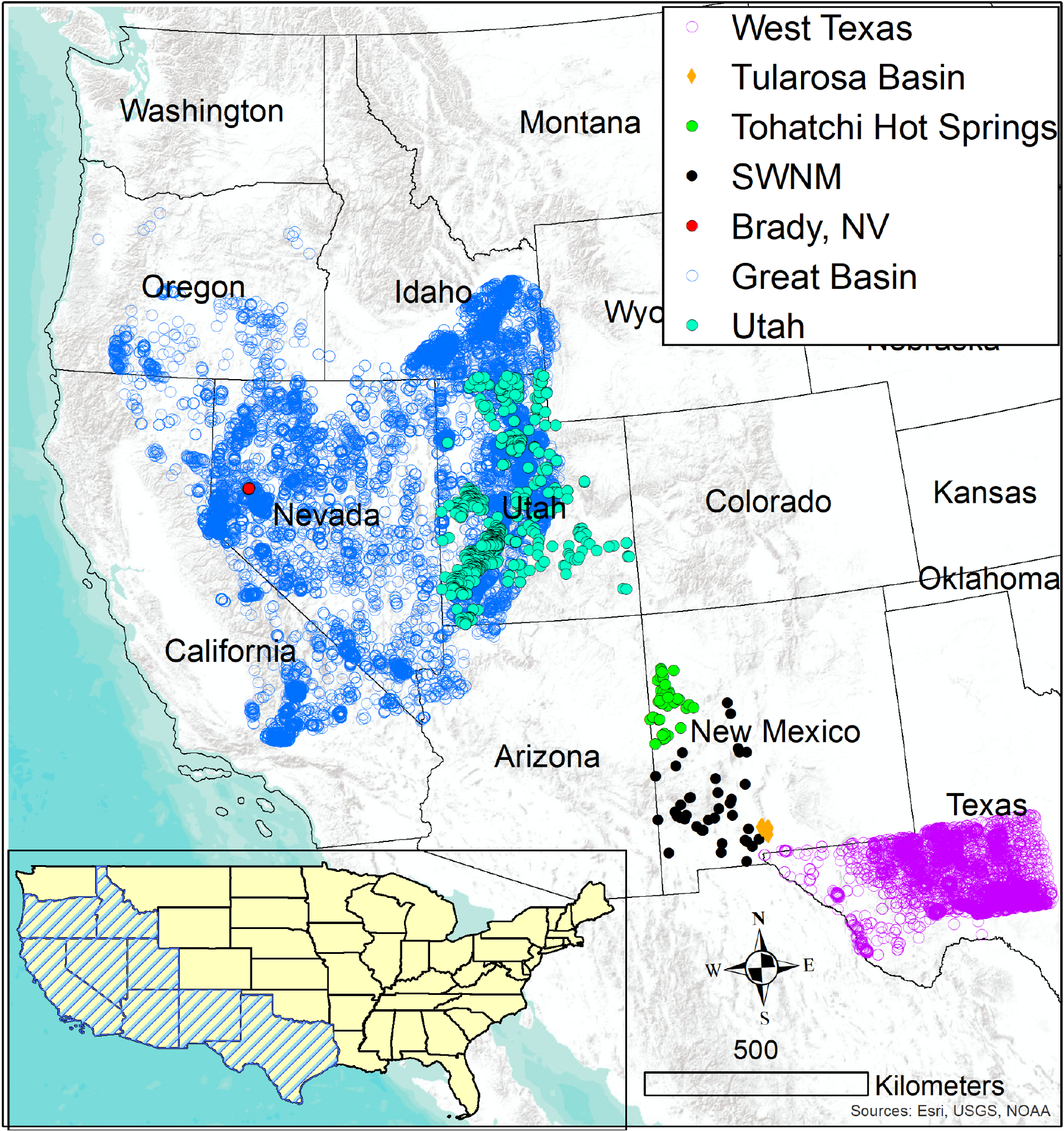}
  \caption{\textbf{Geothermal field data:}~Field locations at which geothermal data is available for ePFA analysis.
  This field data includes thermal, geophysical, geomechanical, geochemical, and geological attributes.
  These datasets are collected and curated at various locations in NV, UT, CA, OR, ID, NM, and HI \cite{}. %PFA GTO link
  \label{Fig:Fig2_pfa_data}}
\end{figure}

%%%%%%%%%%%%%%%%%%%%%%%%%%%%%%%%%%%%%%%%%%%%%%%%%
In this section, we describe the open-source geothermal datasets analyzed using GeothermalCloud.
The GeothermalCloud uses NMF$k$ not only for the execution of ML analyses but also applies the postprocessing and visualization tools provided by NMF$k$.
Datasets include hydrological, geophysical, geomechanical, geochemical, and geological attributes.
The geothermal dataset also covers regional conditions and corresponding resource types (e.g., low, medium, and high temperature) in Nevada, Utah, California, Oregon, Idaho, New Mexico, Texas, and Hawaii Islands in the USA.
Figure \ref{Fig:Fig2_pfa_data} shows the spatial locations of the analyzed data.
All the data attributes are rescaled within the range of 0.0 to 1.0 using a unit range transformation.
If needed, some of the attributes are log-transformed.
In the following paragraphs, we briefly describe the processed data for ML analysis.
Additional information on these datasets are described in references \cite{ahmmed2020unsupervised,ahmmed2020non,vesselinov2021geo}, which were curated from Geothermal Data Repository \cite{weers2016doe,weers2022geothermal}.

%----------------------------------;
% Figure-2: Geothermal field data  ;
%----------------------------------;

The Nevada geothermal PFA Phases I, II, and III identified Great Basin as the largest region with hidden geothermal resources.
The study area comprises of 14,341 spatial locations at which geochemical data is sampled.
A total of 18 different geochemical attributes are available for ML analysis.
Within the UtahFORGE region, 22 data attributes, including satellite (InSAR), geophysical (e.g., gravity, seismic), geochemical, and geothermal attributes, are collected at 102 locations.
Twenty-one data attributes collected during the Phase I and II at 120 different locations are analyzed in the Tularosa Basin.
In the Tohatchi Hot Spring area in Northern New Mexico, we investigated 19 data attributes observed at 41 wells.
A total of 15 data attributes collected at 247 locations during Phase I and II of Hawaiian geothermal PFA are analyzed using NMF$k$.
The data attributes employed in the current ML study are related to geothermal processes.
For instance, they can be used as a proxy or indirect measure of rock-water interactions and infer reservoir temperatures, geothermal conditions, reservoir boundaries, and heat source type (e.g., meteoric, magmatic, mixed) \cite{dobson_review_2016,fridriksson_application_2007,klein_advances_2007,fournier_chemical_1977}.

%%%%%%%%%%%%%%%%%%%%%%%%%%%%%%%%%%%%%%%%%%%%%%%%%
\section{Results and discussion}
\label{sec:results}
%%%%%%%%%%%%%%%%%%%%%%%%%%%%%%%%%%%%%%%%%%%%%%%%%
In this section, we provide the results of our ML methodology on the PFA datasets described in Sec.~\ref{sec:data}.

%==========================;
% Subsection: Great basin  ;
%==========================;
\subsection{Great basin}
\label{subsec:Great_basin}
Great Basin includes multiple known and hidden geothermal resources, which have been studied in the past decades \cite{faulds2016nevada}.
GeoThermalCloud is used to analyze geochemical data, which can indirectly infer geothermal conditions (e.g., reservoir temperatures, conditions, boundaries, and heat source type).
Geochemistry provides insight into water-rock interactions and fluid circulation at various depths (e.g., shallow, deep).
ML analysis is performed on a curated data matrix consisting of 18 data attributes (e.g., temperature, pH, total dissolved solids, anions, and cation concentrations) at 14,321 locations.
Our NMF$k$ algorithm identified three optimal signatures based on the loss function, $\mathcal{L}$, and the Silhouette width value.

\begin{figure}[htbp]
  \centering
    \includegraphics[width = 1.05\textwidth,page=2]{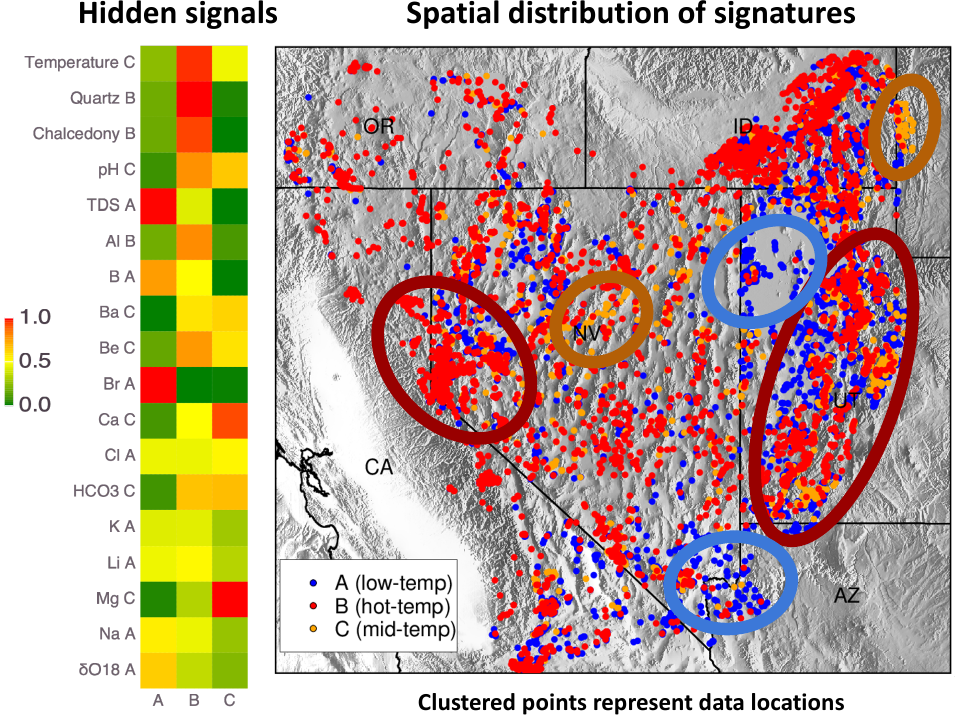}
  \caption{\textbf{ML analysis of Great Basin data:}~ML results for geothermal data within the Great Basin region in Nevada.
  The left figure shows the optimal hidden geothermal signatures discovered by NMF$k$, which is three.
  The right figure provides the spatial distribution of these signatures mapped onto the geothermal data collection locations based on resource type (e.g., modestly, highly, and moderately prospective).
  The ellipses mark the regions with a high density of similar resource-type signature locations.
  \label{Fig:Fig3_pfa_great_basin}}
\end{figure}

Figure~\ref{Fig:Fig3_pfa_great_basin} shows the dominant attributes in the discovered signatures, the spatial distribution of these identified signatures, and their association with a resource type (e.g., low, medium, high).
Signatures A, B, and C define modestly, highly, and moderately prospective hydrothermal systems.
Signature A represents modestly hydrothermal systems because of the low contribution of groundwater temperature in this extracted signature.
The dominant attributes of this signature are TDS, Br$^-$, B$^+$, and $\delta O^{18}$.
Signature B represents highly prospective hydrothermal systems due to the high contribution of temperature.
The dominant attributes of this signature are pH, Al$^{3+}$, Be$^{2+}$, and quartz and chalcedony geothermometers.
Signature C defines moderately prospective hydrothermal systems because of the medium contribution of temperature. The dominant attributes of the signature are Mg$^{2+}$ and Ca$^{2+}$.
Signatures B and C distribution suggest that the significant portions of the Great Basin region have prospective geothermal resources.
Areas with a high density of B and C locations are labeled with ellipses in this Figure~\ref{Fig:Fig3_pfa_great_basin}.
Some of these locations align with existing geothermal resources and sites such as Dixie Valley and Brady geothermal areas in Nevada \cite{faulds2016nevada}.

%--------------------------------------------;
% Figure-3: ML analysis of Great Basin data  ;
%--------------------------------------------;
%%https://www.researchgate.net/publication/348161100_Machine_Learning_on_the_Geochemical_Characteristics_of_Low-_Medium-_and_Hot-temperature_Geothermal_Resources_in_the_Great_Basin_USA
%%https://www.researchgate.net/publication/346962142_Unsupervised_Machine_Learning_to_discover_attributes_that_characterize_low_moderate_and_high-temperature_geothermal_resources
%%https://www.researchgate.net/publication/346582901_MACHINE_LEARNING_TO_CHARACTERIZE_REGIONAL_GEOTHERMAL_RESERVOIRS_IN_THE_WESTERN_USA
%%https://docs.google.com/document/d/1-BciVKLAy-aqj3981EQQ5hz-kGjHJ1TFnq4bIoSy4WM/edit#
%%https://permalink.lanl.gov/object/tr?what=info:lanl-repo/lareport/LA-UR-20-26184
%

%=========================;
% Subsection: UtahFORGE  ;
%=========================;
\subsection{UtahFORGE}
\label{subsec:Utah_FORGE}
\begin{figure}[htbp]
  \centering
    \includegraphics[width = 1.05\textwidth,page=3]{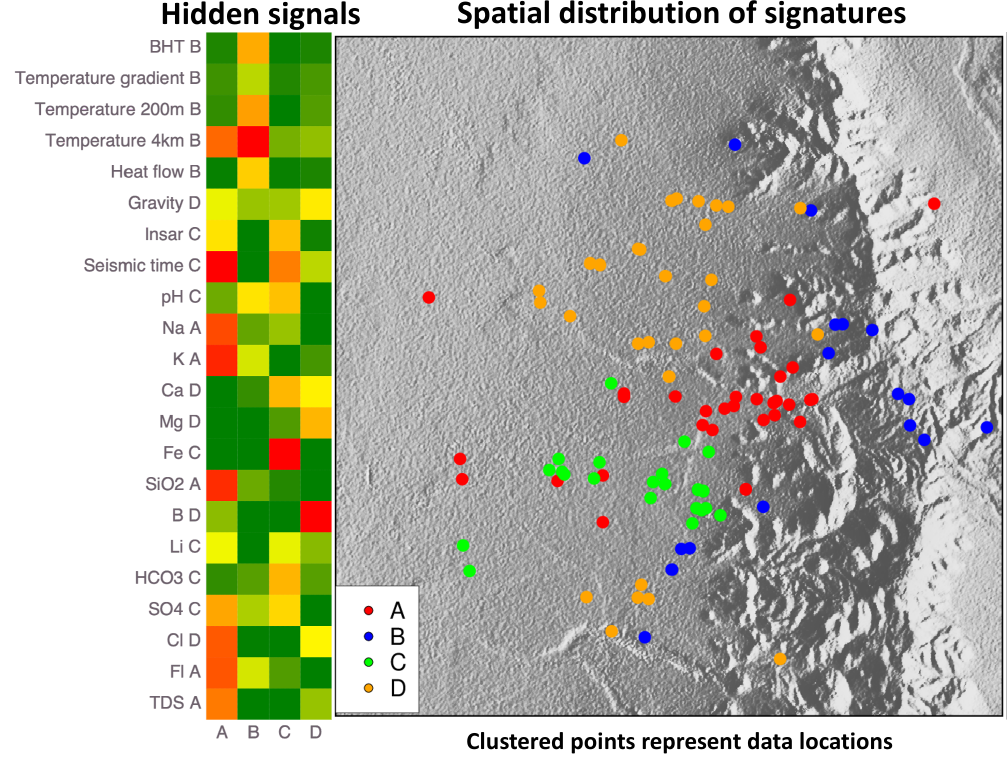}
  \caption{\textbf{ML analysis of UtahFORGE data:}~ML results for UtahFORGE geothermal data.
  The left figure shows the four hidden geothermal signatures discovered by NMF$k$.
  The right figure provides the spatial distribution of these signatures based on low, medium, and high-temperature resource types.
  \label{Fig:Fig4_pfa_utah_forge}}
\end{figure}

UtahFORGE (Frontier Observatory for Research in Geothermal Energy, Utah) is a field laboratory located in Milford, Utah, to study enhanced geothermal systems (EGS) \cite{moore2019utah,allis2016egs}.
This field laboratory provides a unique opportunity to develop and test new technologies for characterizing and creating sustainable geothermal systems in a controlled environment.
During two phases of EGS site development in UtahFORGE, collected geothermal data at 102 locations with 22 attributes.
These include satellite (InSAR), geophysical (gravity, seismic), geochemical, and geothermal attributes.
The NMF$k$ method was applied to this dataset and discovered four optimal hidden geothermal signatures.
Figure~\ref{Fig:Fig4_pfa_utah_forge} shows the discovered signatures, their relationship to data attributes, and the associated mapping of these signatures to the sampled locations.
Signatures A and B are related to favorable geothermal conditions, which have different dominant attributes.
Signature A's key attributes are gravity, seismic, slip rate, and specific geochemical species.
The high contribution from the slip rate indicates that locations covered by Signature A are influenced by EGS activity.
Signature B's dominant attributes are temperature, heat flow, pH, and ions such as K$^{+}$, and Fl$^{-}$.
These heat, pH, and temperature attributes are good indicators of potential resources that may be hidden within this region.
In a nutshell, NMF$k$ analysis solidifies the idea that signature B locations can be a source to discover blind systems in Utah.

%-------------------------------------------;
% Figure-4: ML analysis of UtahFORGE data  ;
%-------------------------------------------;
%https://www.researchgate.net/publication/352454167_Prospectivity_Analyses_of_the_Utah_FORGE_Site_using_Unsupervised_Machine_Learning
%https://www.researchgate.net/publication/344954777_Non-negative_Matrix_Factorization_to_Discover_Dominant_Attributes_in_Utah_FORGE_Data?channel=doi&linkId=5f9af636458515b7cfa9147b&showFulltext=true
%%

%=============================;
% Subsection: Tularosa basin  ;
%=============================;
\subsection{Tularosa basin}
\label{subsec:Tularosa_basin}
\begin{figure}[htbp]
  \centering
    \includegraphics[width = 1.05\textwidth,page=4]{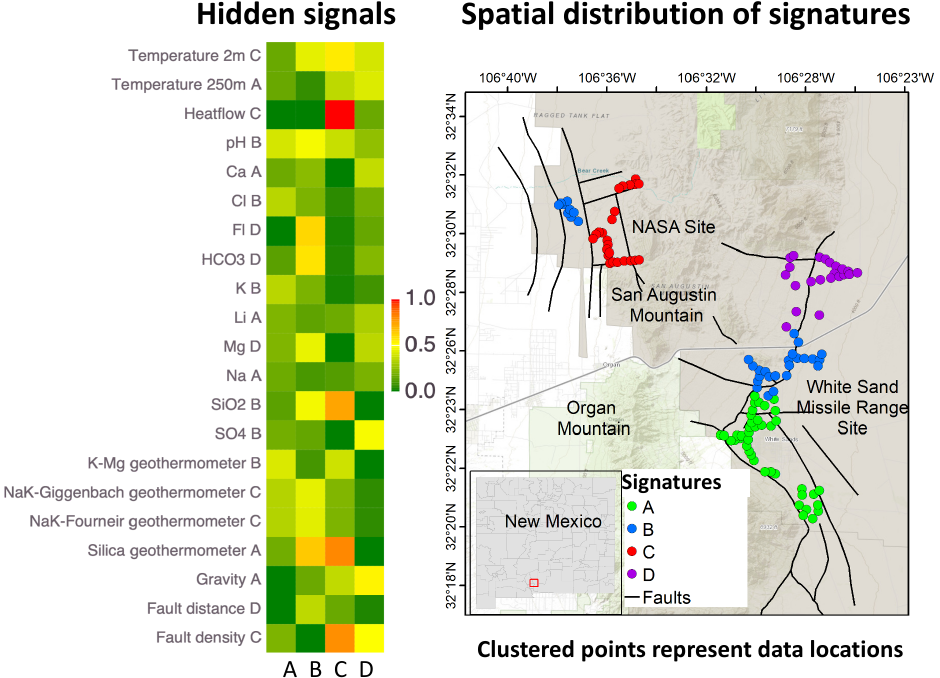}
  \caption{\textbf{ML analysis of Tularosa basin data:}~ML results for geothermal data from the Tularosa Basin in New Mexico.
  The left figure shows the four hidden geothermal signatures discovered by NMF$k$.
  The right figure provides the spatial distribution of these signatures.
  Signature C has dominant attributes (e.g., heat flow, $\mathrm{SiO}_2$, silica geothermometer, fault density) that correspond to the location of hidden geothermal resources in this region.
  \label{Fig:Fig5_pfa_tularosa_basin}}
\end{figure}

The Tularosa basin in the southern Rio Grande Rift within New Mexico consists of promising geothermal plays based on Phase-I and II PFA studies \cite{brandt2015tularosa,nash2017tularosa,nash2017phase}.
The presence of known geothermal systems, high-temperature slim hole wells, quaternary faults, and relatively high heat flow sheds light on possible blind geothermal systems that may be present in the study area.
During the traditional Tularosa basin PFA, various datasets representing the heat of the Earth, such as water chemistry, temperature gradients, 2\,m temperature, and heat flow, are collected to assess the geothermal potential.
We investigate 21 attributes collected during this PFA, including field geology (geological reconnaissance and mapping), gravity surveys, shallow temperature surveys, well water sampling and geothermometry, temperature logging, and magnetotelluric surveys.
Figure~\ref{Fig:Fig5_pfa_tularosa_basin} shows the results of NMF$k$ applied to this Tularosa basin PFA dataset.
Signature C defines the potential for hidden hydrothermal resources due to favorable characteristics associated with geological structures for geothermal exploration.
Critical attributes in this signature include heat flow, $\mathrm{SiO}_2$, silica geothermometer, and fault density.
The spatial distribution of this signature suggests that resources are close to NASA and other military installations, making geothermal energy cost-effective for power needs for several facilities in this region.

%-------------------------------------------;
% Figure-5: ML analysis of Tularosa basin   ;
%-------------------------------------------;

%===================================;
% Subsection: Tohatchi Hot Springs  ;
%===================================;
\subsection{Tohatchi Hot Springs}
\label{subsec:Tohatchi_Hot_Springs}
\begin{figure}[htbp]
  \centering
    \includegraphics[width = 1.05\textwidth,page=5]{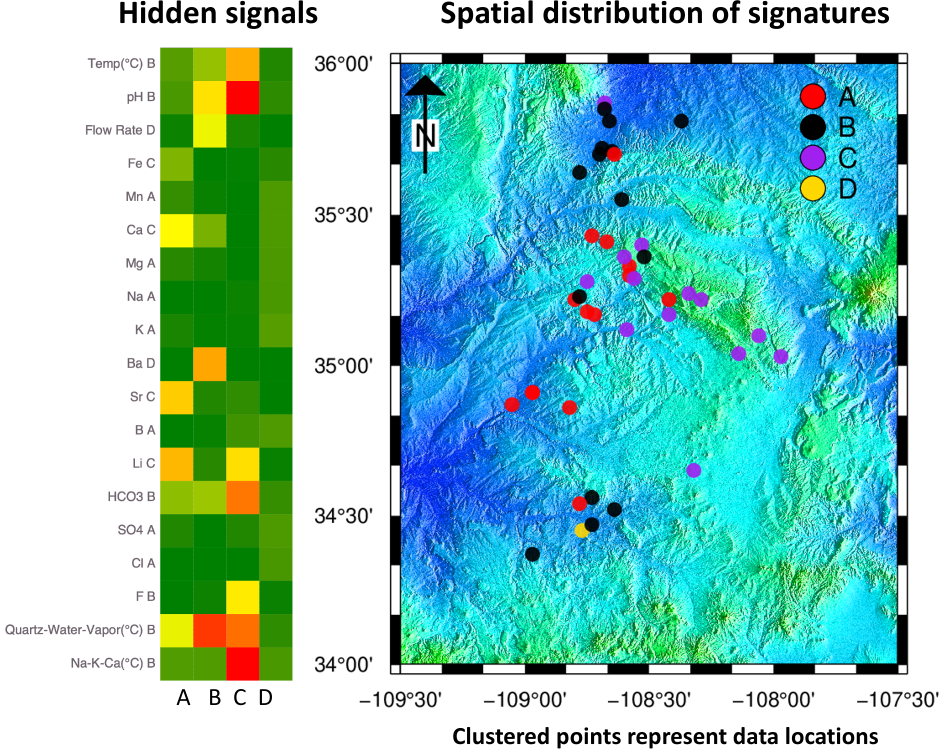}
  \caption{\textbf{ML analysis of Tohatchi Hot Springs data:}~ML results based on geothermal data for the Tohatchi Hot Springs region in New Mexico.
  The left figure shows the four hidden geothermal signatures discovered by NMF$k$.
  The right figure provides the spatial distribution of these signatures, where signature C defines the hidden potential of high-temperature geothermal resources within this region.
  \label{Fig:Fig6_pfa_tohatchi}}
\end{figure}

Tohatchi Hot Springs in New Mexico are favorable for enhanced geothermal exploration \cite{levitte1980geothermal,ahmmed2021geothermal,ahmmed2022exploration}.
Characterizing the geothermal source and governing mechanisms (e.g., identifying which data attributes contribute to the heat source) that make the water hot in these springs can help us better explore this region.
NMF$k$ is a suitable method to understand the resource type better.
We investigated 19 data attributes observed at 41 wells and Tohatchi's hydrogeology to assess if groundwater circulates from a shallow depth (~400 m due to shallow geothermal system, secondary fracture permeability effects) or from a greater depth (e.g., deep circulation of water, forced-convection geothermal systems).
Figure~\ref{Fig:Fig6_pfa_tohatchi} shows the result of our NMF$k$ analysis on the well data, which identified four hidden signatures.
Signature C defines the hidden potential geothermal resources, whose key attributes are pH, Li$^+$, $\mathrm{HCO}_3^-$, F$^+$, Quartz-water-vapor geothermometer, and Na-K-Ca geothermometer.
Our ML analysis of the well data attributes and its signature may suggest that Tohatchi Hot Springs' geothermal reservoir is large for long-term production of hot well water ($<100^{\mathrm{o}}$C)
ML analysis in connection with the SME technical reports allows us to hypothesize that groundwater circulates from a shallow depth due to low geothermal gradients.
%However, certain wells have a high-temperature gradient.
%This may be due to fluid circulation from depth or shallow uranium deposits within that region.
%Further data collection and associated ML analysis are needed to reinforce our hypothesis to efficiently explore geothermal resources within the Tohatchi Hot Springs region.

%-------------------------------------------------;
% Figure-6: ML analysis of Tohatchi Hot Springs   ;
%-------------------------------------------------;
%https://permalink.lanl.gov/object/tr?what=info:lanl-repo/lareport/LA-UR-21-23827
%%

%=============================;
% Subsection: Hawaii islands  ;
%=============================;
\subsection{Hawaii islands}
\label{subsec:Hawaii_islands}
\begin{figure}[htbp]
  \centering
    \includegraphics[width = 1.05\textwidth,page=6]{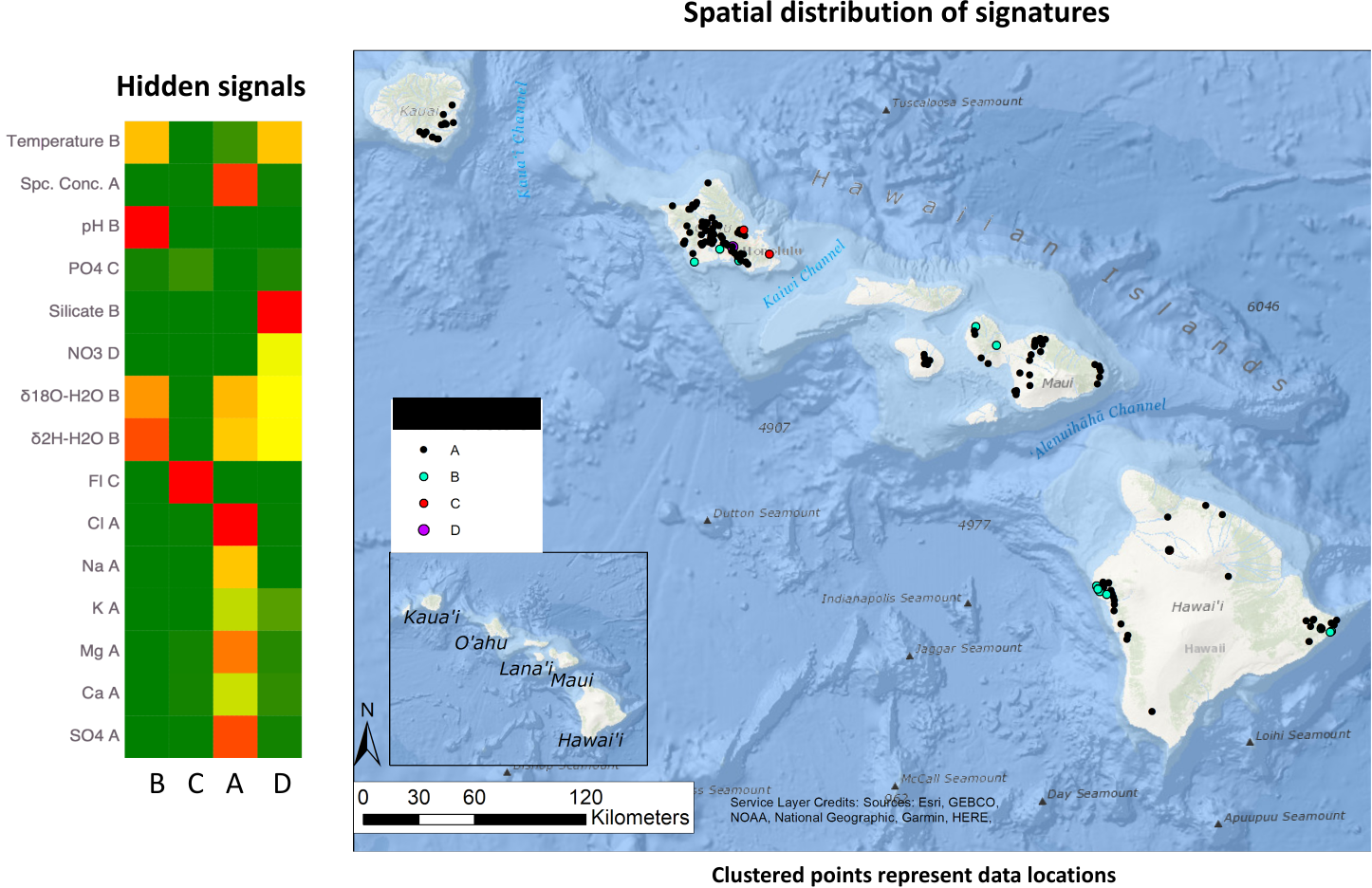}
  \caption{\textbf{ML analysis of Hawaiian islands data:}~ML results based on geothermal data for the Hawaiian islands.
  The left figure shows the four hidden geothermal signatures discovered by NMF$k$.
  Their dominant attributes include pH, $\delta^{18}$O, $\delta^{2}$H, and silicate.
  The right figure provides the spatial distribution of these signatures.
  Signatures B and D define the hidden potential of geothermal resources within this region as they are linked to groundwater temperature and chemistry.
  \label{Fig:Fig7_pfa_hawaii}}
\end{figure}

Hawaii is an ocean island hotspot environment with a magmatic geothermal heat source.
Currently, geothermal resources within this region provide up to 3\% of the state's energy needs.
The PFA analysis conducted across the State of Hawaii by Lautze and co-workers \cite{ito2017play,lautze2017play} showed that these islands have huge geothermal prospects.
We have applied NMF$k$ to discover hidden features in hydrological data linked to Hawaii's geothermal resources.
The discovered signatures and their spatial distribution is shown in Fig.~\ref{Fig:Fig7_pfa_hawaii}.
NMF$k$ identified four signatures in the PFA data that characterized the islands and their resource types within the State of Hawaii.
Each signature has unique physical significance (e.g., water type, alkalinity).
Hidden signatures with high contributions from temperature are essential for ML-enhanced geothermal exploration.
Specifically, signatures B and D define the hidden potential of geothermal resources within this region as they are linked to groundwater temperature and chemistry.
Signature A is related to geochemistry (e.g., cations and anions), another proxy for blind systems.
This extracted signature shows that further data analysis of each island in the State of Hawaii can help us better understand the geothermal resource type and its source (e.g., meteoric, magmatic, or connate).

%----------------------------------------------------;
% Figure-7: ML analysis of Hawaii islands PFA data   ;
%----------------------------------------------------;
%https://www.researchgate.net/publication/346962240_Unsupervised_Machine_Learning_to_Extract_Dominant_Geothermal_Attributes_in_Hawaii_Island_Play_Fairway_Data
%%

%%%%%%%%%%
\section{Conclusions}
\label{sec:Conclusions}
\begin{figure}[htbp]
  \centering
    \includegraphics[width = 1.05\textwidth]{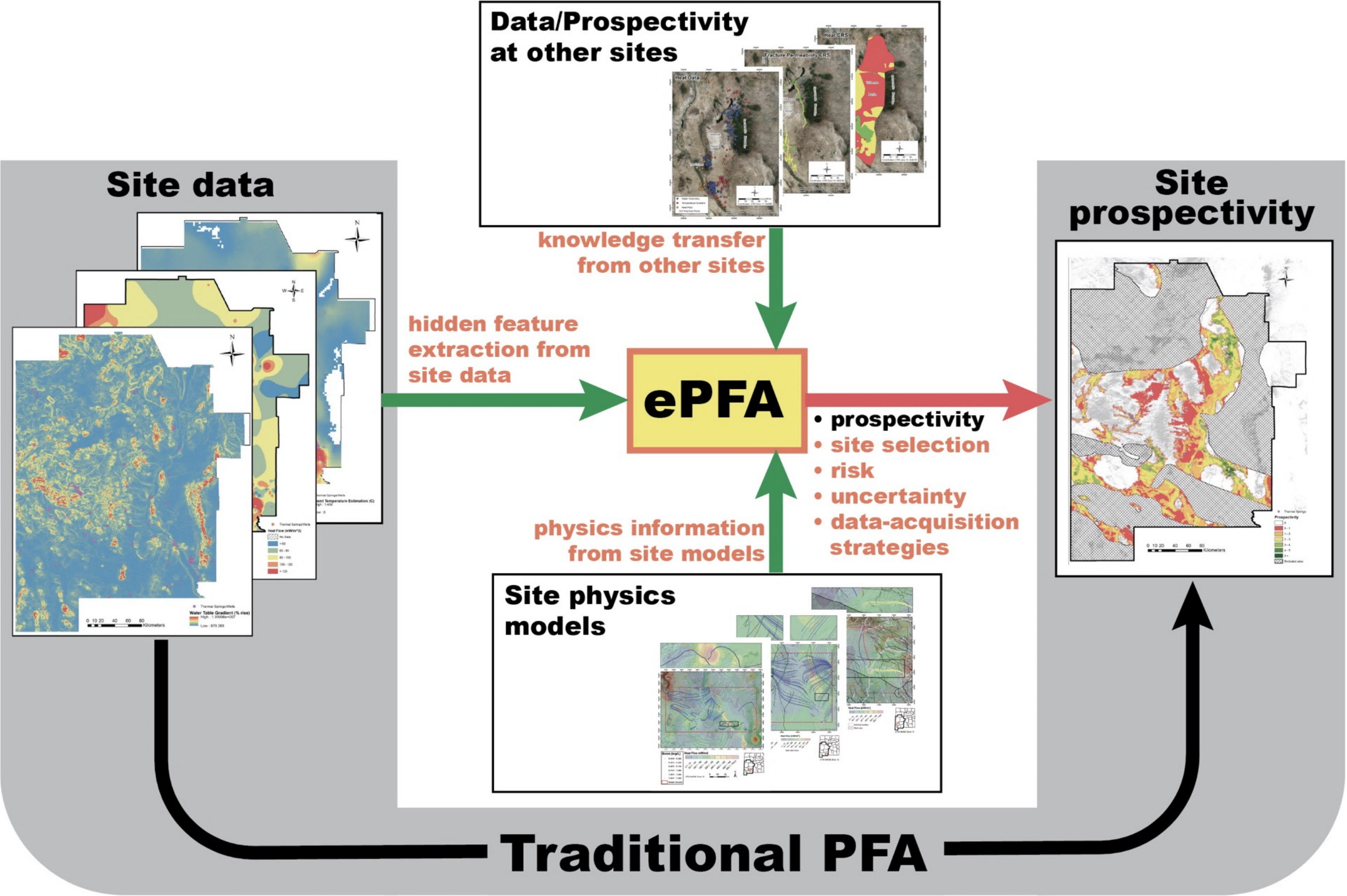}
  \caption{\textbf{Traditional vs. ML-enhanced PFA:}~This conceptual figure compares the traditional and proposed workflow for geothermal PFA using the GeoThermalCloud framework.
  In traditional PFA, geothermal play fairway models are generated using weighted products or sums of geothermal attributes curated by SME's from raw field datasets.
  SME's also perform data weighting in traditional PFA models.
  In ML-enhanced PFA, the final prospectivity maps are generated by combining ML analysis with subject matter expertise.
  ML is used at various stages in the PFA analysis, such as extracting hidden signatures from curated field data, transferring knowledge across sites, and discovering physics information from site-specific process models.
  \label{Fig:Fig1_epfa}}
\end{figure}

The paper presented an ML methodology developed in our open-source framework, GeoThermalCloud \url{https://github.com/SmartTensors/GeoThermalCloud.jl} for efficient geothermal exploration of hidden geothermal resources.
The GeoThermalCloud uses a series of unsupervised, supervised, and physics-informed ML methods in our SmartTensors AI platform.
Here, the presented GeoThermalCloud analyses are performed using our unsupervised ML algorithm called NMF$k$ from the SmartTensors AI platform.
NMF$k$ is applied to analyze and enhance existing public geothermal datasets.
In this way, GeoThermalCloud provides enhanced ML-based exploration results consistent with those obtained from traditional SME-based PFA methods.
The NMF$k$ discovers hidden signatures embedded in the geothermal data, and then SMEs analyze the extracted hidden signatures to understand the discovered knowledge.
We applied the proposed GeoThermalCloud framework to various regions in the U.S. to better characterize the resource types based on the dominant data attributes discovered by NMF$k$.
These regions include the Great Basin in Nevada, UtahFORGE, Tularosa basin, Tohatchi Hot Springs in New Mexico, and Hawaii Islands.
In Great Basin, NMF$k$ identified modestly, highly, and moderately prospective hydrothermal systems, their dominant characterization attributes, and their spatial distribution using geochemistry data \cite{ahmmed2022machine}.
They also compared their results with a comprehensive PFA study conducted in the same study area by Faulds and co-workers \cite{faulds2016nevada}
Fauld et al. used geophysical, geological, geochemical, etc. data yet the prospectivities of two studies are similar with much less data using our ML.
We analyzed site prospectivity within UtahFORGE and potential exploration options (e.g., drilling location for future geothermal field exploration).
In the Tularosa basin, Tohatchi Springs, and Hawaii islands, our NMF$k$ was able to identify low-, medium-, and high-temperature hydrothermal systems and found dominant attributes and spatial distribution for each hydrothermal system.
To conclude, these applications of our GeoThermalCloud for different types of available PFA data instill confidence in our approach to analyze new datasets (e.g., airborne geophysical, 3DEP LiDAR surveys) such as GeoDAWN \cite{rhodes2021analysis}, currently being collected over the parts of Nevada and California.
Figure\ref{Fig:Fig1_epfa} shows a pictorial description of ML-enhanced PFA workflows using our GeoThermalCloud, which we plan to enhance by combining GeoDAWN data with multi-physics simulations.
ML-enhanced GeoDAWN data will provide a deeper understanding of the geothermal conditions (e.g., fault patterns, stress regime) in the Great Basin that represent blind hydrothermal systems, thereby accelerating the discovery of new and undiscovered geothermal resources for domestic use.

%-------------------------------------------;
% Figure-7: Traditional vs. ML-enhanced PFA ;
%-------------------------------------------;

\acknowledgements
This research is based upon work supported by the U.S. Department of Energy's (DOE) Office of Energy Efficiency and Renewable Energy (EERE) under the Geothermal Technology Office (GTO) Machine Learning (ML) for Geothermal Energy funding opportunity, Award Number DE-EE-3.1.8.1.
Los Alamos National Laboratory is operated by Triad National Security, LLC, for the National Nuclear Security Administration of the U.S. Department of Energy (Contract No. 89233218CNA000001).
Pacific Northwest National Laboratory is operated for the DOE by Battelle Memorial Institute under contract DE-AC05-76RL01830.
The views and opinions of authors expressed herein do not necessarily state or reflect those of the United States Government or any agency thereof.

\section*{Nomenclature}
\begin{itemize}
  \item AI:~Artificial Intelligence
  \item DOE:~Department of Energy
  \item EGS:~Enhanced Geothermal Systems
  \item EERE:~Energy Efficiency and Renewable Energy
  \item EGS:~Enhanced Geothermal Systems
  \item ePFA:~ML-enhanced Play Fairway Analysis
  \item FORGE:~Frontier Observatory for Research in Geothermal Energy
  \item GeoDAWN:~Geoscience Data Acquisition for Western Nevada
  \item GTO:~Geothermal Technologies Office
  \item GW:~GigaWatts
  \item InSAR:~Interferometric Synthetic-Aperture Radar
  \item ML:~Machine Learning
  \item NMF:~Non-negative Matrix Factorization
  \item NMF$k$:~NMF + custom $k$-mean clustering
  \item PFA:~Play Fairway Analysis
  \item SME:~Subject Matter Expertise
\end{itemize}

\bibliographystyle{Bibliography_Style}
%\bibliography{Geo_References}
\bibliography{Geo_SmartTensors}

\end{document}